\documentclass[letterpaper, 10 pt, conference]{ieeeconf}  

\usepackage{amsmath,amssymb,amsfonts}
\usepackage{algorithmic}
\usepackage{graphicx}
\usepackage{textcomp}
\usepackage{xcolor}
\usepackage{verbatim}
\usepackage{algorithm2e}
\usepackage{amsmath}
\usepackage{hyperref}
\usepackage[all]{hypcap}
\usepackage{flushend}
\DeclareMathOperator*{\argminA}{arg\,min} 
\bstctlcite{BSTcontrol}

\IEEEoverridecommandlockouts                              

\title{\LARGE \bf Impact Makes a Sound and Sound Makes an Impact: Sound Guides Representations and Explorations}

\author{Xufeng Zhao$^{1, \ast}$, Cornelius Weber$^{1}$,  Muhammad Burhan Hafez$^{1}$, and Stefan Wermter$^{1}$
\thanks{This research was funded by  the German Research Foundation (DFG) in the project Crossmodal Learning (TRR-169) and the China Scholarship Council (CSC).}
\thanks{$^{1}$The authors are with the Knowledge Technology Group, Department of Informatics, Universität Hamburg, 22527 Hamburg, Germany. E-mail: $\{$xufeng.zhao, cornelius.weber, burhan.hafez, stefan.wermter$\}$@uni-hamburg.de.}
\thanks{$^{\ast}$Corresponding author, Email: xufeng.zhao@uni-hamburg.de.}}

\begin{document}

\maketitle
\thispagestyle{empty}
\pagestyle{empty}

\begin{abstract}
Sound is one of the most informative and abundant modalities in the real world while being robust to sense without contacts by small and cheap sensors that can be placed on mobile devices. Although deep learning is capable of extracting information from multiple sensory inputs, there has been little use of sound for the control and learning of robotic actions. For unsupervised reinforcement learning, an agent is expected to actively collect experiences and jointly learn representations and policies in a self-supervised way. We build realistic robotic manipulation scenarios with physics-based sound simulation and propose the Intrinsic Sound Curiosity Module (ISCM). The ISCM provides feedback to a reinforcement learner to learn robust representations and to reward a more efficient exploration behavior. We perform experiments with sound enabled during pre-training and disabled during adaptation, and show that representations learned by ISCM outperform the ones by vision-only baselines and pre-trained policies can accelerate the learning process when applied to downstream tasks.
\end{abstract}

\section{Introduction}
\label{sec:orga0cd83e}
Research in the field of neuroscience shows that with multiple cues from a diverse range of sensory modalities comes enhanced behavioral performance towards faster response, more accurate movement, and a better sense of stimulus \cite{Laurienti04SemanticCongruence}.
When presented with multiple modalities, e.g., a combination of auditory, haptic, and visual perception, an observer will make the \emph{assumption of unity} that decides whether the multimodal information originates from a common source or from some separated objects and events \cite{Welch80ImmediatePerceptual}. The perception of unity arises when the perceiver assumes that a physical event is redundantly expressed and sensed across diverse modalities, and decisions are commonly made based on the temporal and spatial consistency of information \cite{Vatakis07CrossmodalBinding}, or on semantic congruence factors \cite{Laurienti04SemanticCongruence}.

Undoubtedly, vision is extremely information-rich and is one of the most important senses for humans to perceive the world, but is nevertheless hard for a robot to directly extract knowledge from. Though the issue is dramatically alleviated when combined with deep neural networks, visual representations usually are hard to interpret and somehow constrain the tasks they are trained on. For many vision tasks, a common behavior begins by constructing neural networks based on pre-trained models, or by training neural networks in a self-supervised way, e.g., an intra-modal design of simple but diverse sub-tasks \cite{Doersch17MultitaskSelfsupervised}, or crossmodal prediction of information consistency \cite{Zhao18SoundPixels, Arandjelovic18ObjectsThat}. However, only the later design can, at least partially, persist the assumption of unity.

In most scenarios, a vision-based reinforcement learner requires to learn representations and policy jointly
\cite{Laskin21URLBUnsupervised}. Both 
are highly coupled: sufficient and stable representations are essential for policy learning \cite{Burda19LargescaleStudy}; a diverse and near-optimal 
policy is needed to collect samples to learn unbiased representations.
Humans can benefit from multiple sensing cues in terms of both perception and behavior. Intuitively, an active agent who is allowed to explore freely can benefit from multimodal cues in two aspects: 1) learning meaningful representations by crossmodal self-supervision \cite{Eisermann21GeneralizationMultimodal,Higgen20CrossmodalPattern,Parisi18NeuroroboticExperiment}, and 2) being intrinsically motivated to explore the environment under the unity assumption reflected by the uncertainty of crossmodal predictions.

Sounds are generally much more distinctive compared with visual events.
For some specific tasks related to physical properties estimation, the sound alone is reliable to guide a robot and measure its performance \cite{Clarke18LearningAudio}. For others, it may be informative but not sufficient, e.g., a classification of objects that share common auditory properties \cite{Mir21HumanoidRobot}, or precise control of a water-pouring robot \cite{Liang20RobustRobotic}. In this case, sounds are supposed to fuse with other sensory inputs to present a much more robust description of states, or to scaffold the agent's exploration.

There are more chances that sound is abundantly distributed while hardly considered for general manipulations due to the facts that 1) vision is content-rich and is thus sufficient for traditional planning-based robots so the sound is often ignored; 2) the correlation of sound events with a task goal could be implicit to program or to discover automatically by traditional methods, which further limits its exploitation.
However, things go the other way when a deep reinforcement learner is deployed to control. 1) Learning exclusively with vision can be exhausting. Though deep neural networks are capable of extracting features from high dimensional inputs, there is no guarantee of information sufficiency as samples are collected gradually. Representations can possibly overfit the trajectories of a non-optimal agent, especially when transferred to new scenes where a biased policy could lead to a worse learning process. Moreover, exploration time for robots is often desired to be minimal for natural wear and safety concerns, which calls for the necessity of efficient and robust pixel interpretation. 2) Fortunately, latent associations among modalities \cite{Jaegle21PerceiverGeneral,Kumar19StabilizingOffpolicy} and behavior consequences \cite{Silver21RewardEnough} can be discovered automatically by deep learning, which prompts the potential of crossmodal control.

\begin{figure}[htbp]
\centering
\includegraphics[width=.9\linewidth]{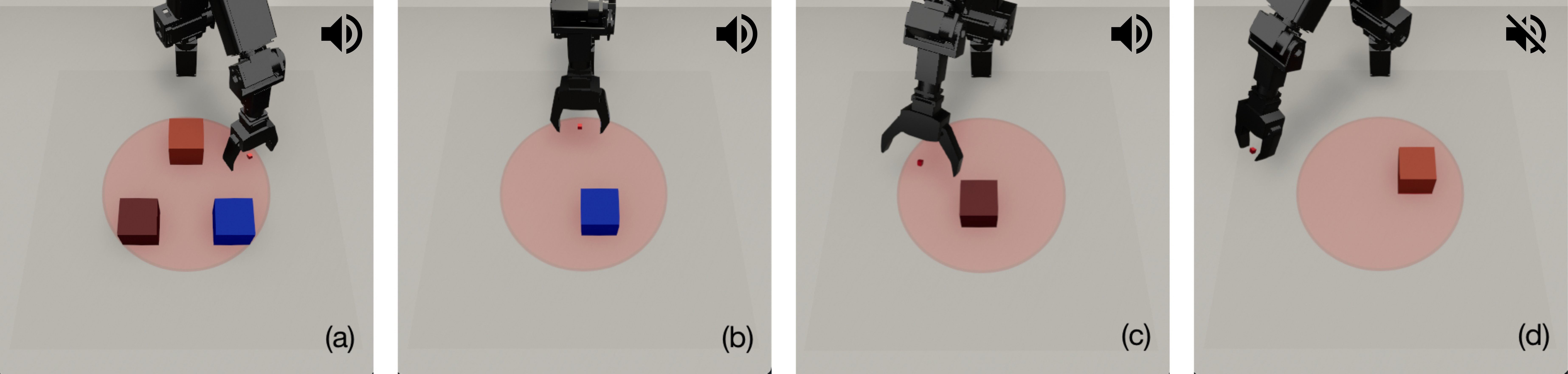}
\caption{\emph{ManipulateSound} environments with different objects that have different physical properties: (a) a task with three different cubes to push out; (b) a fine-tuning task with a single blue ceramic cube to push out; (c) a task with a single brown wooden cube to push out; (d) a task with a single red metal cube to push out; sound intentionally turned off during evaluation.}
\label{fig:env}
\end{figure}

Therefore, our approach contains two phases: first, to train the image encoder of a Reinforcement Learning (RL) agent with visual-auditory correlations, and second, to use the crossmodal error as an intrinsic reward to encourage meaningful exploration.
Contributions in this paper include: 1) the ManipulateSound\footnote[2]{
\url{https://github.com/xf-zhao/ManipulateSound}} environment built upon the ThreeDWorld simulator \cite{Gan21ThreeDWorldPlatform} that comprises robotic control with physically generated sound (see Fig.\ref{fig:env}); 2) a general architecture to utilize sound feedback for unsupervised RL exploration, resulting in more robust representation and active exploration.

\section{Related Work}
\label{sec:org09f01e5}

We introduce sound to boost self-supervised representation learning as well as active exploration of unsupervised RL agents.

\subsection{Self-supervised Representation Learning}
\label{sec:orgdb31ba7}

Self-supervised learning is a collection of methods to learn representations from data that has automatically created pseudo-labels according to certain objectives.
Based on the sensory inputs, they can be roughly classified into two categories: intra-modal and crossmodal self-supervised learning.

A common intra-modal way to create pseudo-labels of images is to perform multiple parameterized augmentations. Then, neural networks are trained to predict which transformation has been carried out on each sample \cite{Doersch17MultitaskSelfsupervised,Agrawal15LearningSee}.
We argue that it makes more sense to a robot when the transformation owns a realistic meaning. For instance, to obtain representations with ego-motion equivariance addressed, images are collected with a camera on a moving car and grouped into neighbor pairs by driving commands \cite{Jayaraman15LearningImage}.
The forward model in the Intrinsic Curiosity Module (ICM) \cite{Pathak17CuriositydrivenExploration} predicts \(s_{t+1}\) with \((s_{t}, a_{t})\) so that the agent can learn to represent the environmental dynamics.

Self-supervised representation learning is naturally applicable for scenarios with multiple modalities involved. Representations emerge concurrently with different focuses and biases, but often have strong relations from one to another. To jointly model multiple modalities, such as audio and visual components of videos \cite{Gao20VisualEchoesSpatial}, a binary classification model to discriminate whether the visual and auditory input are aligned \cite{Arandjelovic17LookListen,Dean20SeeHear}, or a regression model to predict corresponding audio statistics given vision \cite{Owens16AmbientSound} can be established. Although these settings are simple enough, they reveal the unity assumption of events, such that extraordinary abilities can be acquired, e.g., sound localization, audio-visual retrieval \cite{Arandjelovic18ObjectsThat} and 
speech separation \cite{Owens18AudiovisualScene}. In our case, we train a discriminating model which is easy to implement and applicable for general usage.

When applied to robotic control, the available sensory perception is much more diverse \cite{Calandra18MoreFeeling, Murali18LearningGrasp,Gan20LookListen,Chen20SoundspacesAudiovisual}. A work by \cite{Lee19MakingSense} shows that a fused state of visual input, force-torque sensing, and proprioception trained by self-supervision is beneficial for sample efficiency. However, it can be difficult to handcraft such sub-tasks and properly assign weights among modalities. We keep the complexity low by focusing on the impact of sound.

\subsection{Active Exploration}
\label{sec:orgb0cedf4}

A reinforcement learning agent can gain remarkable abilities by purely maximizing the reward of experiences \cite{Silver21RewardEnough}. However, for a task with sparse rewards \cite{Nair18OvercomingExploration, Sekar20PlanningExplore}, which is a common case, the learning process can be quite slow due to the inefficiency of sampling. Reward-shaping \cite{Hu20LearningUtilize} is a commonly used method to alleviate this problem, but it requires expert knowledge and human effort to tune and is 
vulnerable to environmental disturbance.
Many active exploration strategies have been investigated to encourage the agent to seek novel states \cite{Pathak19SelfsupervisedExploration, Eysenbach19DiversityAll,Pathak17CuriositydrivenExploration,Burda19ExplorationRandom} among which ICM proves to be robust on many tasks \cite{Burda19LargescaleStudy,Laskin21URLBUnsupervised}. So we construct our auditory-curiosity module on top of ICM, building on an existing visual processing pathway.

As an alternative to sound, haptic sense \cite{Rajeswar21HapticsbasedCuriosity} achieves good performance and active exploration in terms of frequent contacts, 
supporting sample-efficient learning.
Similar to our work, \cite{Gan20NoisyAgents} use vision and action to predict next clustered auditory events, and the classification error will thus be used as the overall intrinsic reward.
However, the transferability of learned representations is not as well studied as in our work.
Work in \cite{Dean20SeeHear} trains a discriminator to exploit information consistency of aligned image sequences and audio, and intrinsic reward is computed according to the uncertainty of the classifier. Despite the extra efforts required to construct offline data sets, they are restricted to Atari games or audio-dense scenarios. When applied to robotic control, an object will only produce sound when there is a contact. Silence or background noise dominates most of the time. It is even harder to construct misaligned pairs because a random shuffle strategy fails in cases where silence is capable of being aligned with most of the visual scenes. Moreover, a cold-starting problem will arise, particularly when the policy is not sufficiently rewarded to produce collisions. Therefore, we use intrinsic motivations extracted from both visual and auditory cues.

\section{Intrinsic Sound Curiosity Module}

Typical reinforcement learning problems are formulated as Markov Decision Processes (MDPs), comprised by states \(\mathcal{S} = \{ s_{t} \}\), actions \(\mathcal{A} = \{ a_{t} \}\), transition probability \(\mathcal{P}_{ss'}^{a}\), and rewards \(\mathcal{R} = \{ r_{t} \}\).
The goal of the agent is to find the optimal policy \(\pi^{*}(s_{t}, a_{t})\) that maximizes the expected discounted sum of rewards \(\mathbb{E}_{\pi} \sum_{n=0}^{\infty} \gamma^{n} r_{t+n}\).
Usually, out of realistic constraints and generality consideration, we do not have full access to internal states $\mathcal{S}$ but a series of sensors attached to the workspace, resulting in partial observations $\mathcal{O}=\{o_t\}$.
Before being fed into the policy module, high-dimensional sensory inputs must be compressed to latent states that can efficiently represent the environment
\cite{Mnih15HumanlevelControl,Burda19LargescaleStudy}.

\subsection{Visual Representation Learning}
\label{sec:org3f7fe0a}

Visual exploration is a fundamental task for embodied AI agents, where the agent is allowed to actively gather visual information about the environment and then distill knowledge into models such as a topological map or a dynamics model \cite{Duan22SurveyEmbodied}. Generally, the agent is supposed to explore as many novel states as possible with an internal encouragement aligned to certain targets, e.g.\ a measure of the \textit{coverage} such as the amount of visited unique states in a navigation scenario \cite{Dean20SeeHear}, a \textit{prediction error} of a learned dynamics model \cite{Pathak17CuriositydrivenExploration,Du21CuriousRepresentation} or of a reconstruction model when an agent tries to generate other views of an object than the observed ones \cite{Duan22SurveyEmbodied}.

\begin{figure*}[htbp]
\centering
\includegraphics[width=.9\linewidth]{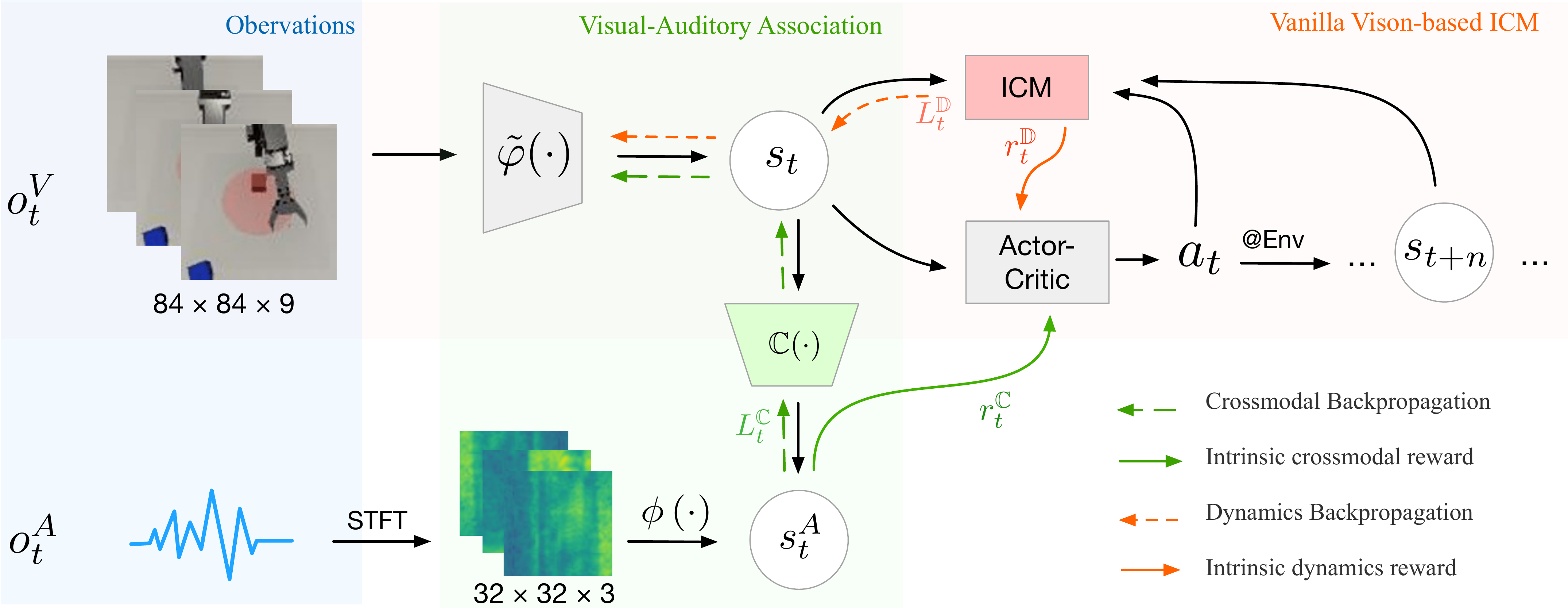}
\caption{\label{fig:iscm}An overview of the Intrinsic Sound Curiosity Module (ISCM) comprised of: 1) visual-auditory observations available in exploration (blue-shaded square), crossmodal learning (green-shaded square) and vanilla vision-based ICM architecture (red-shaded square).}
\end{figure*}

With a combination of multiple sensory inputs for internal states, the agent is allowed to have a more comprehensive view of the environment.
However, it will require either a lot of domain-specific assumptions or an increase in model complexity \cite{Jaegle21PerceiverGeneral,Jaegle22PerceiverIO} to derive efficient representations from fused inputs.
In order to make a fair comparison with vision-only baselines, we use sound in a supplementary way. Only in the pre-training stage has the agent access to sound.
The baseline encoder is trained by dynamically modeling the environment with visual states, while the one of ISCM (Intrinsic Sound Curiosity Module) additionally fits a visual-auditory sub-task (see Fig.\ref{fig:iscm}).
Before adaptation to downstream tasks, visual encoders of the DDPG \cite{Lillicrap16ContinuousControl} learner are initialized with weights from the ISCM and ICM baseline.

Let the visual and auditory observation at time step \(t\) be denoted as \(o_{t}^{V}\) and \(o_{t}^{A}\), respectively.
A visual encoding function \(\varphi(\cdot)\) comprised of convolutional neural networks is thus applied on \(o_{t}^{V}\) to compute the state \(s_{t} = \varphi(o_t^V)\), which is later used for both policy learning and dynamic environment modeling.
Evidence shows that a well-pre-trained encoder is essential for the generalization of supervised learning models \cite{Pan10SurveyTransfer,Doersch17MultitaskSelfsupervised} and RL agents \cite{Laskin21URLBUnsupervised,Burda19LargescaleStudy}.
Hence, the sound-free visual encoder $\varphi(\cdot)$ and the sound-guided counterpart $\tilde{\varphi}(\cdot)$ are trained separately for comparison.

There are two jointly-trained dynamics models in ICM: a forward model \(\mathbb{D}^{F}\) and an inverse model \(\mathbb{D}^{I}\).
The forward model tries to predict the forward n-step transition \(s_{t+n}\) given the current state \(s_{t}\) and action \(a_{t}\), i.e.\ $\hat{s}_{t+n} = \mathbb{D}^{F}(s_t, a_t)$, while the inverse one tries to predict the action taken between aligned states $\hat{a_t} = \mathbb{D}^I(s_{t}, s_{t+n})$, which encourages noise-robust representations \cite{Pathak17CuriositydrivenExploration}. These two dynamics models are optimized concurrently with respect to \(L_{2}\) constraints, defined as \(L_{t}^{F} = \Vert \hat{s}_{t+n} - s_{t+n} \Vert_{2}^{2} \) and \(L_{t}^{I} = \Vert \hat{a}_{t} - a_{t} \Vert_{2}^{2}\). Note that here we use \(L_{2}\) loss also for action predictions since we control the continuous actions of the robot arm, otherwise a cross-entropy loss can be considered for discrete actions.


To benefit from sound, a crossmodal prediction model \(\mathbb{C}(\cdot)\), which can be either a discriminator \(\mathbb{C}^D(\cdot)\) or a regressor \(\mathbb{C}^R(\cdot)\), is then trained to learn the associations of concurrent vision and sound.
It is optimized by minimizing the error between the visual-auditory projection $\hat{s}_t^A = \mathbb{C}(s_t)$ and the targeted latent auditory feature \(s_{t}^A = \phi(o_t^A)\), where $\phi(\cdot)$ is a fixed auditory encoder with output suitable for either discrimination or regression.
Typically, to construct auditory features for regression, $\phi(\cdot)$ consists of randomly initialized neural networks, with no requirements of any further training. These representations are compact, stable, and generally reliable \cite{Burda19LargescaleStudy,Burda19ExplorationRandom}, especially when dealing with impact sound whose information density could be low compared to information in speech. Alternatively, $\phi(\cdot)$ can be chosen as a threshold to distinguish valid event sound from background noise, considering the simplicity and the aforementioned knowledge that even with a simple discriminating task, surprisingly good abilities can be acquired through cross-modal learning \cite{Arandjelovic18ObjectsThat,Dean20SeeHear,Zhao18SoundPixels}.
Much of the time in a manipulation scenario, there is just silence before any valid collision or friction happens.
To avoid the model eagerly collapsing to zero prediction and causing dying neurons \cite{Lu20DyingReLU}, we use weighted cross entropy loss by $\omega$ to amplify the importance of positive samples, i.e.\
\begin{equation}
\label{eq:audioloss}
L_{t}^{\mathbb{C}^D} =  - \omega \cdot s_{t}^{A} \log \hat{s}_{t}^{A} - \left(1 - s_{t}^{A} \right) \log (1-\hat{s}_{t}^{A}).
\end{equation}
For regression, the optimization is similar except for an unweighted \(L_{2}\) loss $L_t^{\mathbb{C}^R} = \Vert \hat{s}_t^A - s_t^A \Vert_2^2$. 

To summarize, the optimal encoders for visual representations in vanilla ICM and the proposed ISCM are separately written as
\begin{equation}
\label{eq:icm}
\varphi^{*} = \argminA_{\varphi} \mathbb{E}_{t} \left[
L_{t}^{\mathbb{D}}
\right]
\end{equation}
and
\begin{equation}
\label{eq:iscm}
\tilde{\varphi}^{*} = \argminA_{\tilde{\varphi}} \mathbb{E}_{t} \left[ (1-\alpha) L_{t}^{\mathbb{D}} + \alpha L_{t}^{\mathbb{C}} \right],
\end{equation}
where $L_{t}^{\mathbb{D}} = \beta L_{t}^{F} + (1-\beta) L_{t}^{I}$ is the overall dynamics loss and $\alpha$, $\beta$ are hyper-parameters to mediate the relative importance between modules.
Note that the objective is expected to be minimized over samples with time stamp $t$. Therefore, it is reasonable to encourage the agent to collect informative samples by injecting the model's prediction error, as a form of intrinsic reward, into the agent's exploration objective.

\subsection{Intrinsic Visual-Auditory Reward}
\label{sec:orgd5e73ae}

Unlike typical supervised learning in which the data is drawn from a stationary distribution, RL agents actively seek samples according to the policy that updates towards reward-weighted maximum likelihood estimation \cite{Peters10RelativeEntropy}.
So when dealing with the sparse-reward case, the intrinsic reward mechanism helps prevent representations to focus too much on non-interesting areas.

The visual-auditory reward in our case is defined as \(r_{t}^{\mathbb{C}} = \log ( L_{t}^{\mathbb{C}} + \epsilon )\) --- if the agent's assumption violates its perception, it will be encouraged to experience more, and vice versa.
$\epsilon$ is a constant added to maintain numerical stability, particularly for values near zero.
With \(r_{t}^{\mathbb{D}} = \log ( L_{t}^{\mathbb{D}} + \epsilon)\) as the ICM reward when modeling the environment dynamics, the overall intrinsic reward of ISCM is computed as
\begin{equation}
\label{eq:overallreward}
r_{t} = \lambda r_{t}^{\mathbb{C}} + (1-\lambda) r_{t}^{\mathbb{D}},
\end{equation}
where \(\lambda\) controls the relative importance of crossmodal prediction and dynamics modeling for exploration.

\subsection{Learning}
\label{sec:orgd30efea}

The learning process is separated into 1) fully unsupervised pre-training and 2) task-specific fine-tuning stages with the curiosity mechanism omitted.
It begins with an agent freely exploring an environment, trajectories of $\{o_t^V\}$ and $\{o_t^A\}$ are accumulated for representation learning; intrinsic rewards are computed for policy learning.
When the freedom limit is reached or when the agent is believed to have enough knowledge, the pre-trained visual encoder will be fixed, and the actor-critic networks will be fine-tuned on downstream tasks with only vision and extrinsic sparse rewards accessible. Refer to Algorithm \ref{alg:iscm} for pseudo code.

\RestyleAlgo{ruled}
\SetKwComment{Comment}{$\triangleright$\ }{}
\begin{algorithm}
\caption{Pseudo Code for ISCM Learning}
\label{alg:iscm}
Initialize: Replay buffer $\mathcal{D} \gets \emptyset$, policy neural networks $\pi$, visual encoder $\tilde{\varphi}$, auditory encoder $\phi$\;
\For(\Comment*[f]{Exploration}){$n = 1$ \KwTo $N_{pre-train}$}{
Observe $o_t = \{o_t^V$, $o_t^A \}$\;
$s_t \gets \tilde{\varphi}(o_t^V)$, $s_t^A \gets \phi(o_t^A)$\;
$a_t \gets \pi(s_t)$\;
Observe $o_{t+1} \sim \mathcal{P}_{ss'}^{a}$\;
$\mathcal{D} \gets \mathcal{D} \cup (o_t, a_t, o_{t+1})$\;
Sample $\mathcal{D}_{batch}$ from $\mathcal{D}$\;
Update $\tilde{\varphi}$, $\pi$ using samples in $\mathcal{D}_{batch}$ with Eq. \ref{eq:iscm} and Eq. \ref{eq:overallreward}\;
}
Fix visual encoder $\tilde{\varphi}^* \gets \tilde{\varphi}$ for evaluations\;
Chose task $T$\;
$\mathcal{D} \gets \emptyset$\;
\For(\Comment*[f]{Adaptation}){$n =1$ \KwTo $N_{fine-tune}$}{
Observe $o_t^V$\;
$s_t \gets \tilde{\varphi}^*(o_t^V)$\;
$a_t \gets \pi(s_t)$\;
Observe $o_{t+1}, r \sim \mathcal{P}_{ss'}^{a}$\;
$\mathcal{D} \gets \mathcal{D} \cup (o_t, a_t, r, o_{t+1})$\;
Sample $\mathcal{D}_{batch}$ from $\mathcal{D}$\;
Update $\pi$ using samples in $\mathcal{D}_{batch}$ with extrinsic rewards\;
}
Evaluate $\pi$ with the accumulated rewards on task $T$ for performance\;
\end{algorithm}
\section{Experiments}
\label{sec:org417c465}

\subsection{Environments}
\label{sec:org123fff1}
The experiments are carried out in simulation because unsupervised exploration in the real world is costly which we leave for future work. One way to manipulate objects with authentic sound is to use a fixed data set with a physics computation interface \cite{Gao21ObjectFolderDataset}.
For generality, we build our manipulation scenarios based on ThreeDWorld (TDW) \cite{Gan21ThreeDWorldPlatform}, a novel embodied AI simulator \cite{Duan22SurveyEmbodied} which is built upon the Unity game engine with multimodal capacities.
To the best of our knowledge, it is the only one so far that supports physically simulated impact and scrape sounds 
\cite{Traer19PerceptuallyInspired,Agarwal21ObjectbasedSynthesis} at run time.
The tabletop robot is composed of a 6-DoF OpenManipulator-Pro robotic arm and a 2-DoF gripper\footnote[3]{\scriptsize \url{https://github.com/ROBOTIS-GIT/open_manipulator_p}}.
It is allowed to manipulate cubes with diverse physical properties that are essential for both dynamics and sound characteristics, e.g., masses, materials, and bounciness.

\textbf{Observations} A camera and a single-channel microphone are placed above the table to capture observations. We focus more on vision and sound, so the robot's proprioception is not included, and the robot has no knowledge of the objects' coordinates. 

\textbf{Rewards} One or several cubes are randomly placed inside a red circular area, and the goal is to push them out of the circle within a limited number of steps.
Specifically, each step will have a penalty of -1/50, and an immediate reward of 1 will be delivered once the task is completed, otherwise the episode ends at 50 steps.

\subsection{Implementations}
\label{sec:orgabe3525}
We use the ICM implementation of URLB \cite{Laskin21URLBUnsupervised} as the baseline, and further extend it to our ISCM architecture\footnote[4]{\scriptsize \url{ https://github.com/xf-zhao/ISCM}}.

\subsubsection{Visual Observation}
\label{sec:org8b17b63}
a) Raw RGB image observations (\(o_{t-2}^{V}\), \(o_{t-1}^{V}\), \(o_{t}^{V}\)) are stacked to the size of $84 \times 84 \times 9$ pixels. b) Four layers of CNN with ReLU activation are applied subsequently to encode vision to a latent state \(s_{t}\). c) A model with two layers of fully connected neural networks with ReLU activation is constructed for sound prediction. d) Visual inputs are available in both pre-training and fine-tuning.

\subsubsection{Auditory Observation}
a) An auditory observation \(o_{t}^{A}\) is generated at run-time by a physical engine; it is then converted to the spectrogram \(o_{t}^{S}\) using Short-Time Fourier Transform (STFT). This is a consideration that complex sounds that come from objects with distinct materials are more distinguishable in the frequency domain with the help of the Fourier transform. Since the agent is updated with samples from a replay buffer and actions are chosen solely based on the visual input, there is no wait for the computation of STFT in real-time control. b) Spectrograms (\(o_{t-2}^{S}\), \(o_{t-1}^{S}\), \(o_{t}^{S}\)) are then stacked as the auditory input of $32 \times 32 \times 3$ size. c) Finally, $s_t^A$ is obtained by applying a certain threshold for silence discrimination; and by passing through a fixed auditory encoder with 36-dimensional output for regression. Auditory inputs are available only in pre-training.

\subsubsection{ICM Modeling}
\label{sec:org0606355}
a) Trajectories of (\(s_{t}\), \(a_{t}\), \(s_{t+n}\)) are fed into  the ICM dynamics models for both encoder training (Eq.\ref{eq:icm} with \(\beta=0.5\)) and intrinsic reward \(r_{t}^{\mathbb{D}}\) computation with $\epsilon=1$. b) The sample with \(r_{t}^{\mathbb{D}}\) is thus used to train a DDPG base learner. c) After enough explorations, the DDPG model will have to adapt to tasks with supervised rewards.

\subsubsection{ISCM Modeling}
a) Paired multimodal observations (\(o_{t}^{V}\), \(o_{t}^{A}\)) are used to train the visual encoder (Eq.\ref{eq:iscm} and Eq.\ref{eq:audioloss} with \(\omega,\alpha,\beta=100, 0.2, 0.5\)) and to compute intrinsic crossmodal rewards \(r_{t}^{\mathbb{C}}\). b) Overall intrinsic reward (Eq.\ref{eq:overallreward} with \(\lambda=0.8, \epsilon=1\)) is thus computed to train a DDPG-based learner.

All the mentioned neural networks are optimized by RAdam \cite{Liu20VarianceAdaptive} with a learning rate equal to 0.001. For many unsupervised RL approaches, the performance decays with an excessive number of environment interactions \cite{Laskin21URLBUnsupervised}.
There is so far no general strategy to determine when to early-stop explorations for better generality.
We empirically choose 200K environment steps to pre-train and 30K steps to fine-tune, considering the convergence of learning curves.
The result is averaged over 4 runs with different seeds.
\subsection{Evaluation}
\label{sec:orgfc29938}
The performance of unsupervised agents can be evaluated by means of measuring the adaptation process on downstream tasks or by statistically analyzing data diversity, e.g., counting of collisions \cite{Gan20NoisyAgents}, variance in the introduced sensory vector \cite{Rajeswar21HapticsbasedCuriosity}, or transformations (distance of movement, orientation changes) of objects.
However, the latter method varies from task to task and is not always applicable.

Whereas the main focus of this work is to demonstrate the effectiveness of learned representations, the tasks are chosen to be simple to master for an agent. In this case, accumulated reward rather than success rate is more appropriate to compare the learning efficiency because the former can reflect the consumed steps, under the setting that the agent is punished for every unfinished step.
Following Michael et al. \cite{Laskin21URLBUnsupervised}, task rewards are solely used as the evaluation metrics: 1) as a measurement of active interactions, extrinsic rewards are accumulated but never leaked during pre-training; 2) the extrinsic rewards accumulated in the adaptation stage.

\subsection{Results and Discussion}
\label{sec:org45f52f6}

\begin{figure}[htbp]
\centering
\includegraphics[width=.99\linewidth]{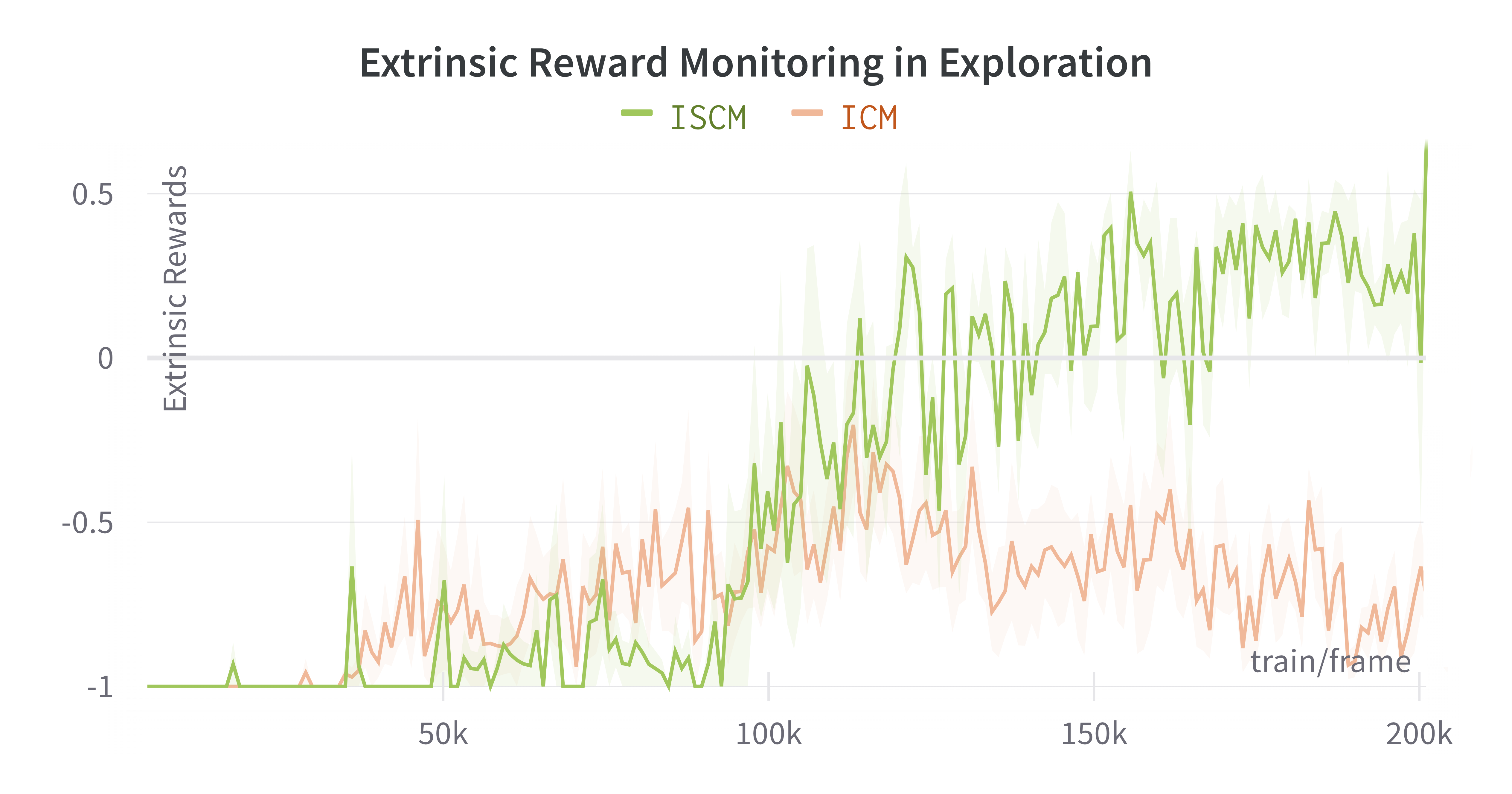}
\caption{\label{fig:extrinsicmonitor}Monitoring of extrinsic rewards (recorded but never used) in exploration. The ISCM agent has more chances of accidentally accumulating extrinsic rewards as a result of sound contributing to additional rewards.}
\end{figure}

We observe that with sound involved, the agent is more interested in interacting with objects, resulting in more occasions of accidental completions (see Fig.\ref{fig:extrinsicmonitor}).

\begin{figure}[htbp]
\centering
\includegraphics[width=.99\linewidth]{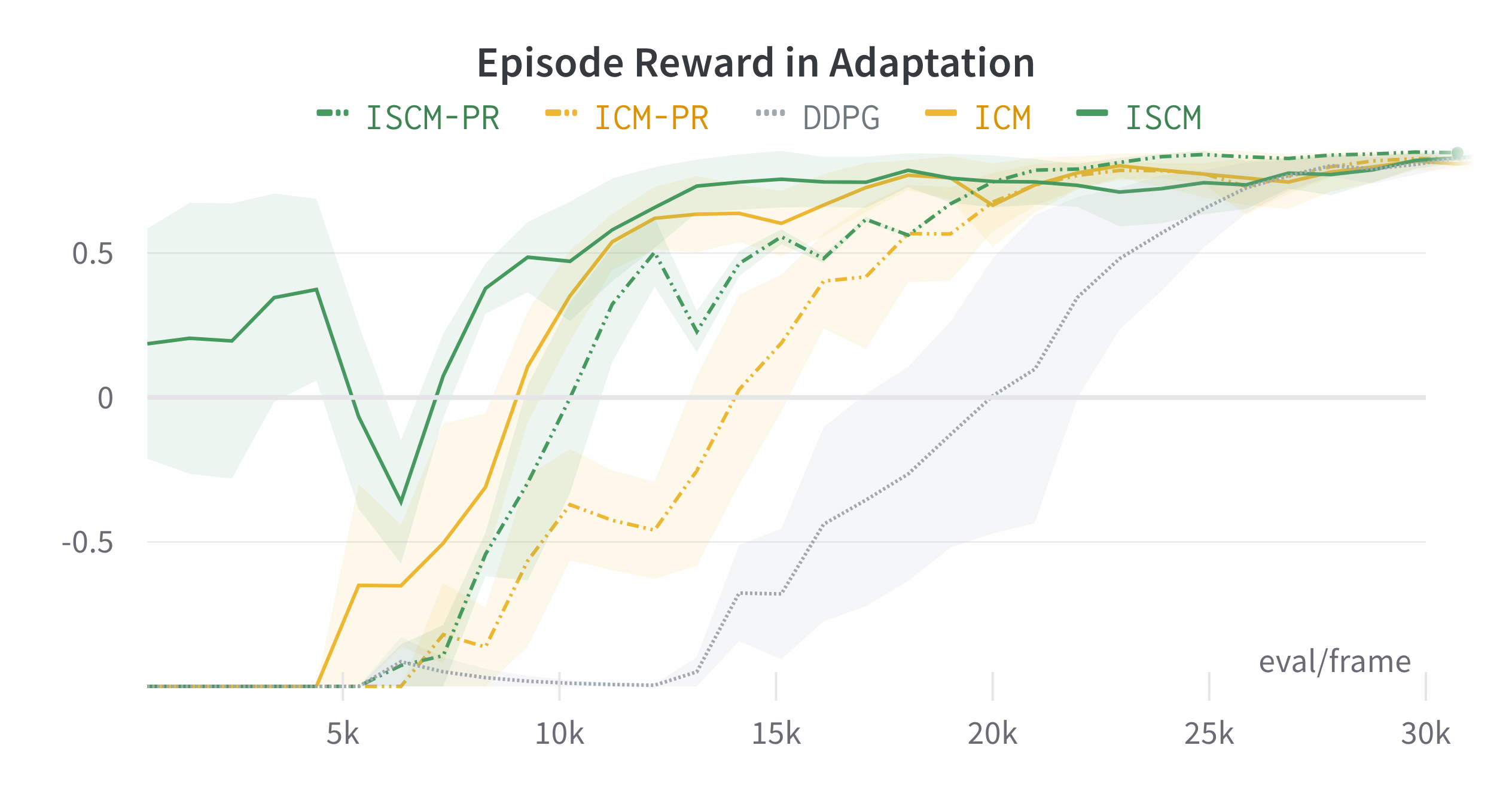}
\caption{\label{fig:soundhelp}Episode rewards in fine-tuning stage accumulated by DDPG learners with all hyper-parameters configured the same except for the initialization of models: 1) ICM: models with representations and policy pre-trained by ICM. 2) ICM-PR: models with ICM pre-trained representations but a re-initialized policy. 3) ISCM: models with representations and policy pre-trained by ISCM. 4) ISCM-PR: models with ISCM pre-trained representations but a re-initialized policy. 5) DDPG: models without pre-training. }
\end{figure}

Observations in the Unsupervised Reinforcement Learning Benchmark (URLB) \cite{Laskin21URLBUnsupervised}
 indicate that the learned representations are universally generalizable while the behavior policy maybe not, especially the policy learned with perfect states (full observable MDPs).
 As is shown in Fig.\ref{fig:soundhelp}, we reiterate that representations learned in unsupervised exploration are essential, and add further findings:
\begin{itemize}
\item There is a big performance gap between the DDPG learned from scratch (DDPG, dashed gray curve) and the other four with pre-trained weights (colored curves), which suggests that unsupervised exploration is helpful for faster adaptation to new tasks.
\item The full pre-trained module (representations and behavior policy) with sound (ISCM, solid green curve) outperforms the baseline that solely depends on vision (ICM, solid orange curve).
\item Without considering pre-trained policies, representations learned with a visual-auditory prediction (ISCM-PR, dashed green curve) outperform the ones learned with only vision (ICM-PR, dashed orange curve).
\item Moreover, by comparing all solid with dashed curves, we find pre-trained policies to have positive effects on task adaptation, which reveals that skills acquired in unsupervised exploration are also reusable. However, more studies on policy analysis, e.g.\ decomposition of the learned policy for abstract behaviors are required for a clear view.
\end{itemize}

\begin{figure}[htbp]
\centering
\includegraphics[width=.99\linewidth]{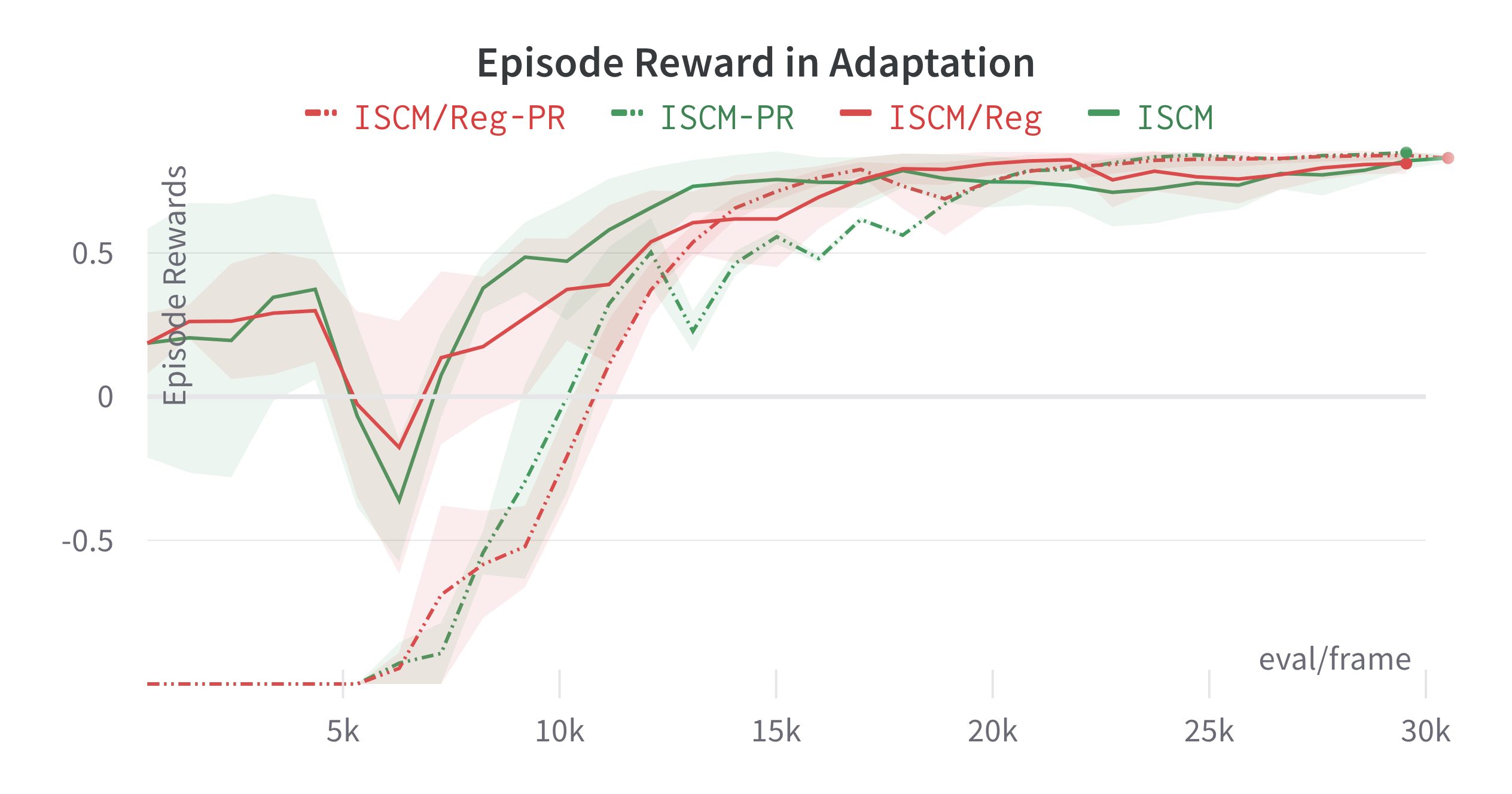}
\caption{\label{fig:rd}Episode rewards in fine-tuning stage are accumulated by base DDPG learners that are initialized differently. 1) ISCM: fully pre-trained module with a discrimination auditory encoder. 2) ISCM/PR: pre-trained representations (but policy re-initialized) with a discrimination auditory encoder. 3) ISCM/Reg: fully pre-trained module with a regression auditory encoder. 4) ISCM/Reg-PR: pre-trained representations (but policy re-initialized) with a regression auditory encoder.}
\end{figure}

A vision-to-sound regression model is also trained with all other hyper-parameters configured the same (see Fig.\ref{fig:rd} for a clear comparison). Though a vector (for regression) rather than a scalar (for discrimination) is believed to have a higher capacity of information, we find the discriminator setup (green curves) achieves a comparative performance with a regressor (red curves), while being simple to implement. Similar findings can be also found in recent works \cite{Gan20NoisyAgents} where clustered auditory events are being predicted instead of regressing sound features. It may be a result of the following reasons: 1) impact sound presents not much more information than a deduction of event occurrence; 2) simulated sound is still far away from perfect such that vision, sound, and dynamics are not matched well as in reality. Future work will include constructions of more complex environments and sim-to-real adaptations to investigate more on these research questions.

\section{Conclusions}
\label{sec:org9011a71}

Sound is one of the most common and efficient modalities, but is yet less considered to learn either simulated or real-world robotic manipulations.
Unlike many of the curiosity-driven RL variants, especially the ones combined with audio that pay attention to non-robotics applications such as playing Atari games, we are focusing on investigating how robots can benefit from exploring multimodal environments.
In this paper, the importance of unsupervised representation learning and of active exploration is addressed.
We further propose the ISCM architecture to use physics-based sound as guidance regarding both aspects.
Our experiments demonstrate that a sound-guided reinforcement learner is more active and has a great superiority to form sufficient as well as stable representations over vision-only baselines.
In future work towards more applicable scenarios, we anticipate novel and interesting robot behaviors emerging in multimodal environments.

\bibliographystyle{IEEEtran}
\bibliography{IEEEabrv,root.bib}

\end{document}